\documentclass[letterpaper]{article} % DO NOT CHANGE THIS
\usepackage{aaai2027}  % DO NOT CHANGE THIS
\nocopyright
\usepackage[hyphens]{url}  % DO NOT CHANGE THIS
\usepackage{graphicx} % DO NOT CHANGE THIS
\usepackage{natbib}  % DO NOT CHANGE THIS AND DO NOT ADD ANY OPTIONS TO IT
\usepackage{caption} % DO NOT CHANGE THIS AND DO NOT ADD ANY OPTIONS TO IT
\usepackage{algorithm}
\usepackage{algorithmic}

\usepackage{newfloat}
\usepackage{listings}
\DeclareCaptionStyle{ruled}{labelfont=normalfont,labelsep=colon,strut=off} % DO NOT CHANGE THIS
\floatstyle{ruled}
\newfloat{listing}{tb}{lst}{}
\floatname{listing}{Listing}

\usepackage{amsmath}
\usepackage{graphicx}
\usepackage{optidef}
\usepackage{amssymb}
\usepackage{bm}
\usepackage{subcaption}
\usepackage{svg}
\usepackage{placeins}
\usepackage{booktabs}

\title{Value-oriented policy distillation for interpretable wind farm power control}
\author{
   Max Bruninx, Timothy Verstraeten, Jan Decuyper, Ann Nowé, Jan Helsen
}
\affiliations{
    Vrije Universiteit Brussel, Brussels, Belgium
}

\begin{document}

\maketitle

\begin{abstract}
 Wind farm power control can help the system maintain balance and avoid blackouts. In Europe, this can be achieved through \textit{implicit balancing}, where a wind farm adjusts its power setpoint in anticipation of the imbalance price using information published by the Transmission System Operator. Thereby, the wind farm helps restore system balance while also reducing its own imbalance costs resulting from forecast errors. However, implicit balancing is a challenging problem since the settlement price is unknown in advance and can be impacted by the pursued control strategy. Reinforcement learning methods could help to solve this problem, but lack the interpretability needed to be deployed on real wind farms. Therefore, we propose a policy distillation framework using explainable boosting machines (EBMs), a variant of generalized additive models, to obtain a transparent control policy. To increase the value captured by the EBM model, we introduce a value-oriented perspective to policy distillation inspired by research in the field of energy forecasting. Furthermore, we suggest to include expert knowledge during model design, such that the control strategy is also justifiable to stakeholders. Using real data of the Belgian grid and an offshore wind farm, we find that our value-oriented and expert-based controller can reduce imbalance costs by $82\%$, while outperforming a price threshold controller as well as a policy distillation baseline using decision trees.
\end{abstract}


\section{Introduction}

% general intro
Electricity supply and demand must be equal at all times to avoid blackouts. This has, however, become more and more challenging as renewable generation capacity continues to grow \citep{IEA_GlobalEnergyReview2025}. Therefore, the Transmission System Operators (TSOs), who are responsible for grid stability, rely on flexible assets that can help counteract system imbalances. The growing need for flexibility has driven a significant increase in battery storage projects, but controllable renewable assets can also contribute to this \citep{EliaAdequacyFlexibility2025}. In Europe, one way to provide flexibility as a renewable asset is by allowing the power setpoint to deviate from its trading schedule in response to real-time information published by the TSO, a practice known as \textit{implicit balancing}. The producer is then rewarded or penalized for this deviation during imbalance settlement, depending on whether it contributed to restoring system balance. Renewable assets can participate in implicit balancing by curtailing their power output in case the system has a surplus. This practice could reduce imbalance costs resulting from forecast errors and improve their overall revenue. Finding these new revenue streams is particularly important for offshore wind farms, since subsidies have recently been decreasing (see e.g., \citet{totalenergies2024negative, rvo2024negative}).

Implicit balancing is, however, a challenging control problem, since the settlement price is only known after the settlement period and a setpoint that negatively affects system imbalance can result in serious financial losses. Furthermore, this study focuses on offshore wind farms, whose large installed capacity relative to typical system imbalances means that their actions can influence the imbalance price. Reinforcement learning (RL), and actor-critic methods in particular, have been proposed to solve this problem and were found to achieve good performance \citep{madahi2024distributional, ally2025day, madahi2025model, rasic2025safe, bruninx2026every}. Unfortunately, these RL-based controllers are rarely deployed in practice due to their black-box nature. To facilitate deployment on an actual wind farm, a controller must demonstrate interpretability, which comprises two aspects: First, the model should be \textit{transparent}, meaning that the effect of the input variables on the model or control policy should be clear. Second, the input variables should have a clear definition and the actions should be \textit{justifiable} within the domain. The latter implies that a market expert must be able to explain each decision to stakeholders such as the TSO and the wind farm operator. To achieve interpretability, the model or control policy should be developed in interaction with experts from the field.

The goal of our work is to explore whether we can build an interpretable controller based on the control policy learned with soft actor-critic (SAC). To this end, we propose a policy distillation framework using explainable boosting machines (EBMs) \citep{lou2013accurate, nori2019interpretml}, a variant of generalized additive models (GAMs), which were shown to be transparent and attain high predictive power. To the best of our knowledge, we are the first paper to apply EBMs in this context. Moreover, two improvements are made to the EBM model. First, to achieve interpretability, we develop an expert-based model in which the univariate and interaction effects are selected based on the explanations from the complete model combined with expert input on the drivers of implicit balancing. Second, to emphasize the high-value actions during policy distillation, we introduce \textit{value-oriented policy distillation}, where the distillation loss is weighted based on the return associated with each state-action pair.

In our experiments, the proposed method is applied to Belgian data for the years 2022 and 2023. The results indicate that our value-oriented and expert-based control policy can reduce imbalance costs by $82\%$. While the controller was not deployed on a real wind farm, we took a lot of care to make the experiments close to reality. More specifically, we used real measurements from the supervisory control and data acquisition (SCADA) system of Norther, an offshore wind farm in the Belgian offshore zone. To account for the price impact of the control strategy, we have implemented the calculation method used by the Belgian TSO, which makes it possible to derive the price for different scenarios using historical data.

\section{Related work}

% \subsection{Interpretable RL}

In our work, we follow the definition of interpretability by \citet{glanois2024survey}, which makes the important distinction against explainability in the field of artificial intelligence. Explainability relies on an external method or proxy to explain the predictions made by a given model. Examples are post-hoc explanations using techniques such as SHAP \citep{lundberg2017unified} or permutation feature importance \citep{altmann2010permutation}. In contrast, interpretability requires transparency to be an intrinsic property of the model, by using techniques such as decision trees or linear models and avoiding complex input variables. However, in addition to \citet{glanois2024survey}, we argue that interpretability also requires the model to be justifiable within the domain. For example, we can have a transparent model where the effect of an input variable cannot be explained by a domain expert to other stakeholders, then this effect is not justifiable and should be removed.

In the context of RL, interpretability can be achieved by immediately employing transparent models when learning the RL policy, using, e.g., tree-based methods \citep{likmeta2020combining, silva2021encoding} or genetic programming \citep{hein2018interpretable}. Alternatively, one can first learn an RL policy using a non-interpretable technique and then apply policy distillation \citep{rusu2015policy} with a transparent model, which is the approach we follow in this study. To this end, deterministic decision trees were employed by \citet{liu2018toward} and soft decision trees by \citet{coppens2019distilling}. Although soft decision trees are generally considered transparent, they only produce filters for each node which are difficult to explain to end users. Therefore, in  \citet{gokhale2024distill2explain}, the authors map the soft decision trees to deterministic rules at inference. On the contrary, this work adopts EBMs, which are transparent by design and require no postprocessing.

Within policy distillation for interpretable RL, another distinction can be made between value-based and policy-based distillation. Examples of the former are \citet{liu2018toward, gokhale2024distill2explain}, where the aim is to replicate the Q-function learned by a deep Q-network. In contrast, the objective of the latter is to directly distill the explicit policy learned by the RL algorithm (see, e.g., \citet{coppens2019distilling}). In our work, we pursue the second approach. Nonetheless, drawing inspiration from the field of value-oriented forecasting \citep{zhang2022cost, wang2023risk, smets2025value}, we propose to include a value perspective while distilling the explicit policy as explained in Section \ref{sec:value_oriented}.

\section{Soft actor-critic}

\subsection{Formulation as a Markov Decision Process}

To enable applying RL methods to this problem setting, we first formulate the implicit balancing problem as a Markov Decision Process (MDP) using a tuple $\left(\mathcal{S},\mathcal{A},\mathcal{P}, \mathcal{R}, \gamma\right)$. The state and action spaces are denoted by $\mathcal{S}$ and $\mathcal{A}$, respectively. The reward function is defined as $\mathcal{R}: \mathcal{S} \times \mathcal{A} \rightarrow \mathbb{R}$ and the transition function as $\mathcal{P}: \mathcal{S} \times \mathcal{A} \times \mathcal{S} \rightarrow \left[0,1\right]$. Last, $\gamma \in \left(0,1\right]$ denotes the discount factor for future rewards.

To transition towards the next state, we have built a simulator that models the dynamics of the wind farm and the reserve market. First, we apply the power setpoint $x_t$ to the wind farm and assess the actual power production $P_t$. During this step, we account for the possible power $\overline{P}_t$\footnote{The maximum power the wind farm can produce given the wind conditions.} as well as other constraints on the power setpoints imposed by the wind farm operator. We then compute the residual system imbalance, which includes the difference between the power production $P_t$ and the trading schedule $v_t$ of the wind farm, and use it to determine which reserves should be activated to resolve the imbalance. To this end, we employ a similar set-up to \citet{bruninx2026herd} but apply the actual price formula used in Belgium \citep{elia2024brp}. Furthermore, we also include the effect of the balancing strategy pursued by the TSO \citep{haaberg2016classification} using historical data on proactive reserve actions.

Given that there is a publication delay for the information published by the TSO\footnote{This publication delay is 2 minutes in Belgium.}, we are dealing with a partially observable MDP. Therefore, for each timestep $t$, the state space $\mathcal{S}$ is defined as:
\begin{itemize}
    \item The latest published imbalance price estimate: $\hat{\lambda}^{\mathrm{B}}_t$.
    \item A forecast for the system imbalance during the imbalance settlement period (ISP) based on the minute-level system imbalances: $\hat{SI}^{\mathrm{ISP}}_t$.
    \item The accumulated position during the settlement period up to time $t$: $\Delta E_t = \sum_{i\in \mathrm{ISP}, i<t} (P_i - v_i)\Delta t$.
    % = \sum_{i\in ISP, i<t} (P_i - s_i)\Delta t$ with $P_t$ the power production of the wind farm at time $t$.
    \item The estimated power forecast error: $\hat{\varepsilon}_t = \overline{P}_{t-1}-v_t$, which approximates the power forecast error using the possible power at the previous timestep.
    \item The minute of the settlement period: $t^{\mathrm{ISP}} \in [0,14]$.
\end{itemize}

The actions are defined as the deviation of the setpoint $x_t$ from the trading schedule $v_t$: 
\begin{equation}
    a_t = x_t -v_t \in \mathcal{A},
\end{equation}
with a continuous action space $\mathcal{A}\subset\mathbb{R}$.

The reward is obtained at the end of every settlement period and is given by:
\begin{equation}
    r_t = \lambda^{\mathrm{B}}\sum_{i\in \mathrm{ISP}} (P_i - v_i)\Delta t,
\end{equation}
where $\lambda^{\mathrm{B}}$ denotes the settlement price.

% episode lenght = 15min
In contrast to batteries participating in implicit balancing (see, e.g., \citet{madahi2024distributional, madahi2025model, rasic2025safe}), wind farm power production cannot be stored for later use. We therefore set the episode length to match the settlement period (15 minutes). The discount factor $\gamma$ is set to $1$, since every minute within the settlement period contributes equally to the reward received at its end.

\subsection{Learning the optimal policy}
The optimal policy is obtained using soft actor-critic \citep{haarnoja2018soft}, an off-policy deep RL method that employs an entropy term to encourage exploration. The policy is parameterized by an actor network $\pi_\phi$, which maps states to a distribution over actions, and a critic network $Q_\theta$, which estimates the expected return of state-action pairs and guides the actor's updates. To stabilize training, SAC employs twin critic networks and updates their target networks $Q_{\theta'}$ via Polyak averaging.

At each training step, a minibatch of transitions $(s,a,r,s^{\prime})$ is sampled from the replay buffer $\mathcal{D}$, with $a^{\prime}$ denoting the action related to the next state $s^{\prime}$. The actor is trained to maximize expected return while maintaining high entropy, giving the loss
\begin{equation}
    J_\pi(\phi) = \mathbb{E}_{s \sim \mathcal{D}, a \sim \pi_\phi}[\alpha \ln \pi_\phi(a|s) - Q_\theta(s,a)],
\end{equation}
where $\alpha$ denotes the temperature parameter, which weighs the entropy term against the Q-value. The critic is trained by minimizing the soft Bellman residual
\begin{equation}
    J_Q(\theta) = \mathbb{E}_{(s,a)\sim\mathcal{D}}[(y - Q_\theta(s,a))^2],
\end{equation}
with target
\begin{equation}
    y = r +\gamma \mathbb{E}_{s^{\prime} \sim \mathcal{D},a^{\prime}\sim\pi_\phi}\left[Q_{\theta^{\prime}}\left(s^{\prime},a^{\prime}\right) - \alpha \ln \pi_\phi\left(a^{\prime}|s^{\prime}\right)\right].
\end{equation}
Finally, following \citet{haarnoja2018soft_appl}, the temperature parameter $\alpha$ is not fixed but learned via gradient descent on
\begin{equation}
    J(\alpha) = \mathbb{E}_{s \sim \mathcal{D}, a \sim \pi_\phi}\left[-\alpha\left(\ln \pi_\phi(a|s) +\mathcal{H}\right)\right],
\end{equation}
where $\mathcal{H}$ is the target entropy.

\section{Explainable boosting machines}

Explainable boosting machines \citep{nori2019interpretml} are generalized additive models which also include pairwise interaction terms ($\text{GA}^2\text{M}$) \citep{lou2013accurate}. For regression, the models have the form
\begin{equation}
    \mathbb{E}\left[y\right] = \beta_0 + \sum f_i\left(z_i\right) + \sum f_{i,j}\left(z_i, z_j\right), 
\end{equation}
where $\beta_0$ denotes the intercept and $z_i$ the input features. Unlike traditional GAMs, which fit the shape functions $f_i$ using splines, EBMs learn each function by combining bagging and cyclic gradient boosting of shallow decision trees.

The models with pairwise interactions have proven more powerful than traditional GAMs~\citep{hastie1986generalized}, which only model univariate effects, while remaining transparent. Specifically, the relationship between the response variable and the input variables can be visualized through a 2D-plot (univariate effects) or a heatmap (interaction effects). Moreover, for each prediction, the final outcome is obtained as the sum of all the terms, allowing for local explanations.

In a $\text{GA}^2\text{M}$ model, all univariate effects and the most important interaction effects, selected using the FAST algorithm \citep{lou2013accurate}, are included. While this results in an accurate and transparent model, the model explanations are not necessarily justifiable based on the domain knowledge of an expert. Therefore, we propose to first train a full model to achieve the best possible accuracy and then prune or extend the model based on expert input. By doing so, the final model can have a reduced accuracy but will be interpretable. 

% For instance, in our case, the main driver of the control policy should be the intermediate prices published by the TSO as they reflect the system needs \citep{elia2025flexibility} and the potential reward. Therefore, the expert-based policy was constructed using this variable as the only univariate effect. Interaction effects were then included based on the remaining variables of the state space. An example is the interaction between the published price and the moment of publication, which was used to incorporate the effect of the publication delay and how it affects the accuracy of the published price.

\section{Value-oriented policy distillation}
\label{sec:value_oriented}

The goal of policy distillation is to transfer the knowledge from a teacher agent $\pi^T$ to a student agent $\pi^S$ \citep{rusu2015policy}. To this end, the teacher agent is first used to generate a dataset $\mathcal{D}^T=\langle\left(s_i, a_i\right)\rangle_{i=1}^M$. Afterwards, the student agent can be trained to replicate the policy of the teacher. In a continuous action setting, this can be done by minimizing the loss function
\begin{equation}
    L\left(\pi^T, \pi^S\right) = \frac{1}{M} \sum_{i=1}^M\left(a_i-\hat{a}_i\right)^2,
\end{equation}
where $a_i$ and $\hat{a}_i$ are the actions given state $s_i$ for, respectively, the teacher and the student agent.

However, directly distilling the policy of the teacher agent does not make a distinction between high and low value actions within the policy. Nonetheless, we can extend the dataset previously generated to include the return associated with each state-action pair $g=\langle g_i\rangle_{i=1}^M$ and extend the loss function to
\begin{equation}
    L\left(\pi^T, \pi^S\right) = \frac{1}{M} \sum_{i=1}^M \omega\left(g_i\right) \left(a_i-\hat{a}_i\right)^2,
\end{equation}
where $\omega: \mathbb{R}\to\mathbb{R}^+$ is a function which maps the returns to a desired weight depending on the problem setting. By doing so, the effective training distribution is shifted toward higher-value actions. Since, in our setting, $\gamma=1$ and the episode length is set equal to the settlement period, $g_i$ equals the imbalance settlement payment during the period of that timestep. Note that, alternatively, the value weights could be obtained based on the Q-value predictions of the critic network instead of the observed returns.

% In our setting, the agent can receive either a reward or a penalty during imbalance settlement, meaning that the episodic returns can be either positive or negative. Therefore, we first normalize them to the range $\left[-1,\ 1\right]$ and then apply exponential weighting:
% \begin{equation}
%     L_V'\left(\pi^T, \pi^S\right) = \mathbb{E}\left[e^{k\tilde{g}}\left(\pi^T\left(a|s\right)-\pi^S\left(a|s\right)\right)^2\right],
% \end{equation}
% where $\tilde{g}$ is the normalized return and $k>0$ is a scaling factor.

% ,
% where the teacher model was used to generate a dataset $\mathcal{D}^T=\left\{(s_i, \pi^T\left(a|s_i\right))\right\}_{i=0}^N$

\section{Numerical experiments}

\subsection{Set-up}

Experiments are conducted using Belgian data of the years 2022 and 2023. The data comes from the Open Data Platform by Elia \citep{eliaopendata} and the SCADA system of Norther, a Belgian offshore wind farm with an installed capacity of 370 MW. In our set-up, we assume the wind farm offers its day-ahead power forecast to the market instead of bidding strategically (see e.g., \citet{heiser2025betting, bruninx2025day, heiser2026and}). The power forecasts are obtained using the approach described in \citet{bruninx2027probabilistic}, while the system imbalance forecasts are produced with Chronos-2 \citep{ansari2025chronos}. All other variables can be derived from the data directly.

The SAC and interpretable controllers were trained using the first year and a half, whereas the last 6 months were used for evaluation. As baselines, we include decision trees (DTs), with a tree depth of 2 and 3, and a price threshold controller, which is often used in industry. The details of the price threshold controller are described in Appendix \ref{app:price_threshold}. We also evaluated soft decision trees \citep{frosst2017distilling} as an alternative baseline. However, they did not perform competitively in our setting (cf., Appendix \ref{app:sdt}) and were therefore excluded from the main experiments. The SAC agents were trained for $200$k episodes using a learning rate of $3\times10^{-4}$ for the actor network and a learning rate of $1\times10^{-3}$ for the critic network and the optimizer of the temperature parameter $\alpha$. The actor and the critic network are multi-layer perceptrons with two hidden layers containing $256$ neurons each. Furthermore, the soft update factor $\tau$ is set to $0.01$, the replay buffer has a size of $10\times10^{6}$ and the batch size equals $1024$.

For the expert-based EBM model, the univariate and interaction effects were decided based on the justifiable effects of the full EBM model. For instance, in the full EBM model, a positive accumulated position $\Delta E_t$ increased the power setpoint and vice versa. When the controller is actively curtailing, this behavior is justifiable, as it corresponds to a continuation of the existing strategy. However, the accumulated position is driven not only by the control actions but also by the power forecast error. This formulation would therefore imply that, in the absence of curtailment, the strategy is to persist in the direction of the power forecast error irrespective of the system's needs, which is undesired. The term was therefore replaced with an interaction effect between the accumulated position and the estimated power forecast error $\hat{\varepsilon}_t$, ensuring that the setpoint only responds to the accumulated position when the wind farm is in fact curtailed. All effects included in the expert-based model can be found in Appendix \ref{app:expert_based}.

For the expert-based, value-oriented model, we employ the additional profit of the policy relative to the price threshold controller instead of using the observed returns directly. The reason for this being that there are certain states in which any controller will achieve a negative reward. This is, for example, the case when the day-ahead forecast overestimated the possible power production and the settlement price is positive. Here the controller cannot do much to offset the imbalance penalty apart from producing at possible power. To stabilize training, outliers are first capped and the resulting values are normalized to the range $[-1,1]$. Afterwards, we apply exponential weighting to ensure that each observation receives a strictly positive weight. 

\subsection{Performance of the interpretable models}

\begin{table}[t!]
\caption{Performance evaluation metrics across the interpretable models and a LightGBM baseline. Mean absolute error (MAE) and root mean squared error (RMSE) are normalized by the capacity of the wind farm. Lower scores indicate a better fit for MAE and RMSE, while higher scores are better for $R^2$.}
\label{table:model_performance}
\centering
\resizebox{0.75\linewidth}{!}{
\begin{tabular}{lccc}
    \toprule
    & MAE & RMSE & $R^2$ \\
    \midrule
    DT (depth=2) & 0.21 & 0.29 & 0.38 \\
    DT (depth=3) & 0.21 & 0.27 & 0.45 \\
    EBM (full) & 0.18 & 0.23 & 0.61\\
    EBM (expert) & 0.19 & 0.24 & 0.57 \\
    EBM (expert + value) & 0.19 & 0.24 & 0.57 \\
    % \noalign{\vskip 0.75ex}
    % \hdashline
    % \noalign{\vskip 0.75ex}
    \cmidrule(lr){1-4}
    LightGBM & 0.15 & 0.19& 0.72\\
    \bottomrule
\end{tabular}
}
\end{table}

\begin{table}[t!]
\caption{Observed imbalance costs (+) or revenues (-) for the different policies over the evaluation period. The results are presented in € per MW installed per settlement period.}
\label{table:model_pnl}
\centering
\resizebox{\linewidth}{!}{
\begin{tabular}{lccccc}
    \toprule
    & Mean & Min & Max & P05 & P95 \\
    \midrule
    No control & 3.22 & -73.01 & 211.77 & -6.55 & 22.13 \\
    Price threshold & 1.47 & -129.73 & 197.53 & -8.39 & 16.30\\
    % \noalign{\vskip 0.75ex}
    % \hdashline
    % \noalign{\vskip 0.75ex}
    \cmidrule(lr){1-6}
    DT (depth=2) & 1.21 & -105.74 & 193.24 & -9.72 & 16.01 \\
    DT (depth=3) & 0.75 & -116.04 & 193.24 & -13.17 & 15.74 \\
    EBM (full) & 0.36 & -99.00 & 193.24 & -14.13 & 15.56 \\
    EBM (expert) & 0.63 & -73.01 & 193.24 & -13.89 & 16.30\\
    EBM (expert + value) & 0.58 & -73.01 & 193.24 & -13.92 & 16.24 \\
    % \noalign{\vskip 0.75ex}
    % \hdashline
    % \noalign{\vskip 0.75ex}
    \cmidrule(lr){1-6}
    SAC & 0.21 & -114.27 & 195.85 & -15.30 & 15.90\\
    \bottomrule
\end{tabular}
}
\end{table}

In Table \ref{table:model_performance}, we show how accurately each method can distill the SAC policy. We also include the performance of a LightGBM model \citep{ke2017lightgbm} to allow for comparison against a black-box technique that can capture complex interaction effects. Overall, we note that all EBM models outperform the DT baselines, but they still attain $R^2$ values that are rather low. Nonetheless, if we consider the performance of the LightGBM model, we notice that even for a black-box model it is complicated to completely distill the policy learned by the SAC actor and, therefore, the performance is found to be acceptable.

In Table \ref{table:model_pnl}, the performance of the different models is shown in terms of imbalance costs per settlement period. The imbalance costs are determined by the forecast error and the pursued control strategy. All controllers observe a positive cost over the evaluation period, meaning that they were not able to completely offset the effect of the forecast errors. The SAC model achieves a cost reduction of 93\% relative to the scenario without control, whereas the interpretable controllers attain a lower reduction. Nonetheless, they still offer a substantial improvement over the price threshold baseline: the expert-based, value-oriented policy achieves an imbalance cost reduction of 82\% compared to 54\% for the price threshold controller. When comparing the DT and EBM policies, the latter were found to achieve lower imbalance costs. Even the worst-performing EBM attains a 16\% improvement over the best-performing DT. Furthermore, we observe that the full EBM model outperforms the expert-based models. However, the full EBM model includes five univariate and ten interaction effects, which makes predictions more difficult to interpret. Therefore, this decrease in performance is the price paid for an increase in interpretability. Finally, we note that the improvement in performance using value-oriented policy distillation is relatively small. Nevertheless, a decrease in imbalance costs of almost 8\% does imply a significant cost reduction in absolute terms, especially for a model with the same level of interpretability.

\begin{figure}[t!]
    \centering
    \includegraphics[width=0.95\linewidth]{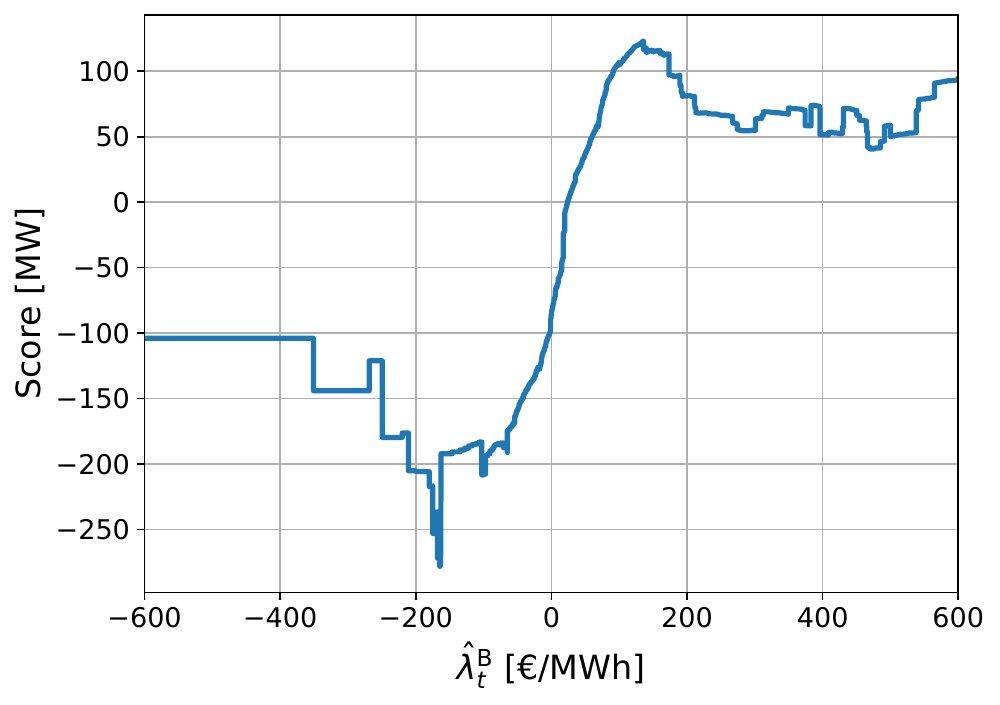}
    \caption{Visualization of the learned shape function for the latest published imbalance price estimate $\hat{\lambda}^{\mathrm{B}}_t$. The score illustrates the effect the variable has on the final prediction. A negative score implies that this variable reduces the power setpoint and vice versa.}
    \label{fig:global_univariate}
\end{figure}

\subsection{Model explanations}

In this section we will show how global and local explanations can be obtained from the EBM models. Global explanations describe how the model works in general, whereas local explanations are used to explain an individual prediction. To obtain global explanations, we can visualize the shape functions learned for each individual and interaction effect. To explain an individual prediction, we show how the output of each shape function contributes to the final control action.

In Figure \ref{fig:global_univariate}, the global effect of the latest published imbalance price estimate $\hat{\lambda}^{\mathrm{B}}_t$ on the control policy is shown. We observe that the policy is in line with what we would expect, since a negative price results in a curtailment action, whereas a positive price results in a power setpoint above the trading schedule. However, interestingly, we also note that for slightly positive prices the controller still decides to curtail. This might be explained by the fact that sometimes flexibility providers offer decremental energy to the TSO at a positive price, meaning that they pay the TSO to be activated, so the imbalance price can be positive when the system has a power surplus. They do so because they also account for the marginal cost of producing the electricity sold beforehand, for example, the fuel consumption cost for a gas turbine. Nonetheless, the capacity offered at this positive price is limited and once it is consumed the imbalance price will become negative, which the controller could be anticipating. Surprisingly, we observe a sharp decrease of the setpoint within the range of $-150$ €/MWh and $-200$ €/MWh compared to the surrounding intervals, which is also true for the SAC policy. Because we expected the control policy to respond more strongly to more negative prices, we investigated this further and found that, within this interval, the final settlement price is more likely to end up negative. A possible explanation may be that other market participants are also reacting to this price using a price threshold controller with a threshold set below this range. Since these participants have not reacted yet, there is a lower chance of the combined reaction of all participants overshooting.

\begin{figure}[t!]
    \centering
    \includegraphics[width=0.95\linewidth]{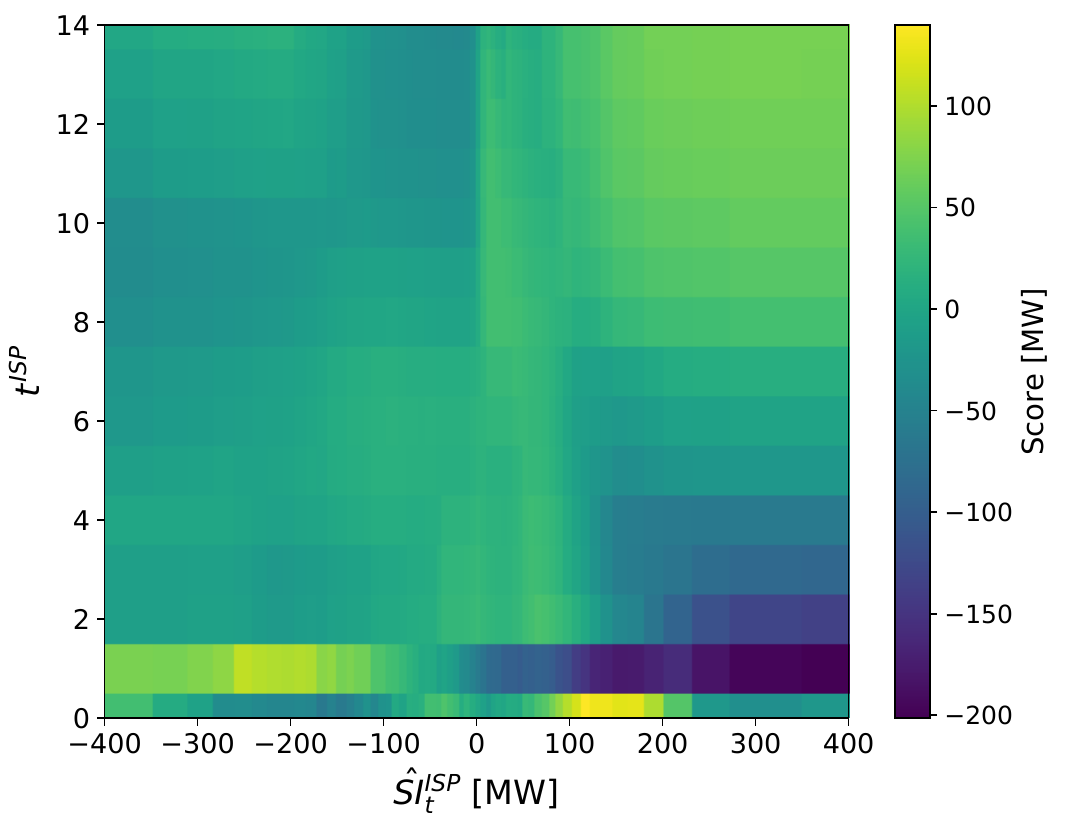}
    \caption{Visualization of the learned shape function for the interaction effect between the system imbalance forecast $\hat{SI}^{\mathrm{ISP}}_t$ and the minute of the settlement period $t^{\mathrm{ISP}}$. A negative score implies that this interaction effect reduces the power setpoint and vice versa.}
    \label{fig:global_interaction}
\end{figure}

In Figure \ref{fig:global_interaction}, we show an example of one of the interaction effects of the model, namely the interaction between the system imbalance forecast and the minute of the settlement period. A first remarkable observation is that the effect for the second minute during the settlement period is clearly different from the effects for the other minutes. Interestingly, this minute is also the last timestep where the imbalance price estimate is still based on the previous settlement period due to the publication delay. This implies that the controller could potentially have learned to rely on the system imbalance forecast, which is forward-looking, because the imbalance price estimate is not reflective of the situation during the current settlement period. Secondly, we observe how the controller is accounting for the price impact of its control strategy. Specifically, we notice that during the first half of the settlement period, a system imbalance forecast around zero results in a small reaction in the same direction as the system imbalance. During the second half of the settlement period, this reaction becomes more extreme as an attempt to not switch the sign of the system imbalance as this would result in a penalty for previously taken control actions in the correct direction.

\begin{figure*}[t!]
    \centering
    \includegraphics[width=0.95\linewidth]{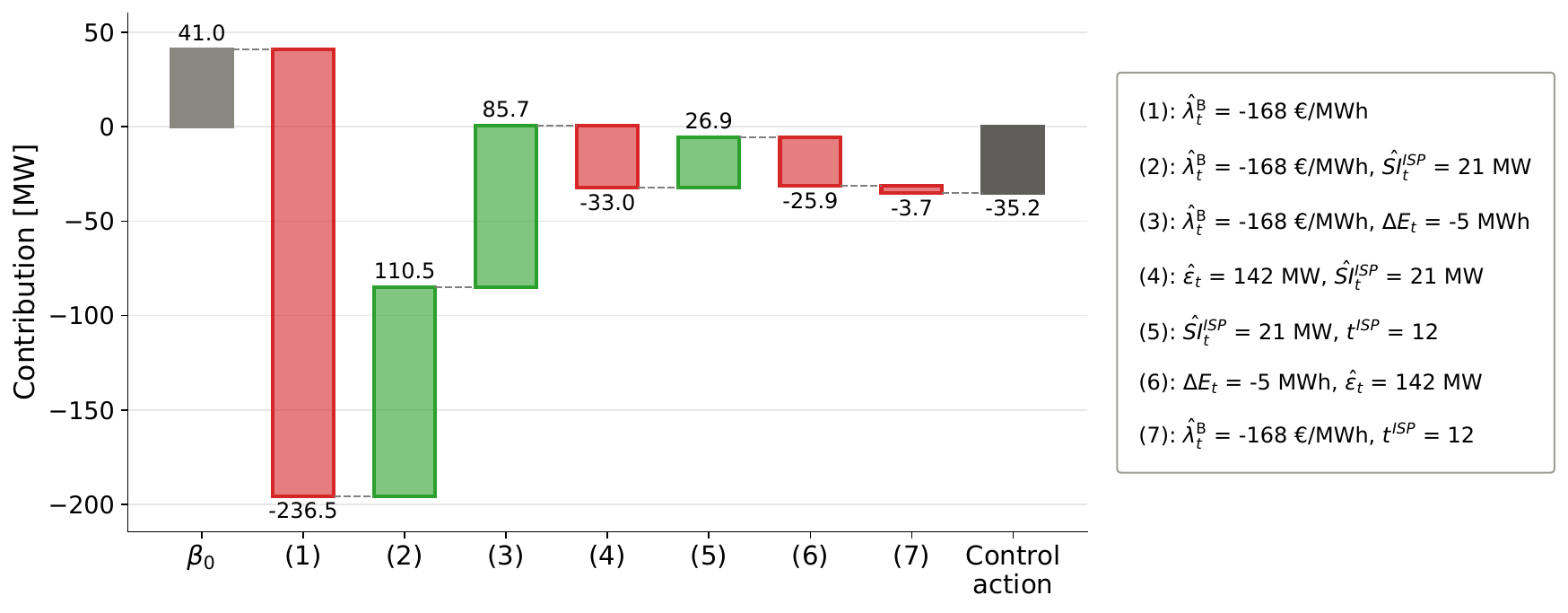}
    \caption{Example of local explanation of the interpretable policy visualized as a waterfall chart. The control action in this example is to reduce the power setpoint by $35.2$ MW compared to the trading schedule.}
    \label{fig:example_ebm}
\end{figure*}

In Figure \ref{fig:example_ebm}, we show an example of a local explanation of a control action that reduces the setpoint compared to the trading schedule. The action could be explained to a stakeholder as follows: (1) The main driver of this action is the fact that we observe a negative price in the range of $-150$~€/MWh and $-200$~€/MWh. (2) However, since the system imbalance only forecasts a moderate surplus for this negative price, the setpoint is increased again. (3) Furthermore, the controller is already expecting a reward for this settlement period and therefore it increases its setpoint to reduce the risk of overreacting. (4) Nonetheless, we observe that both the system and the wind farm have a power surplus and hence the wind farm can curtail to help the system, resulting in reducing the setpoint again. (5) Since the end of the settlement period is approaching and the system imbalance is moderate, the setpoint is slightly increased.  (6) Finally, the controller continues the strategy pursued throughout the settlement period and (7) slightly lowers the setpoint in response to a negative imbalance price estimate toward the end of the settlement period. %While this explanation has been produced manually, we believe it could be automated by combining text templating with large language models and would be an interesting direction for future research.

\section{Conclusion}
\label{sec:towards_deploy}

In this work, we have developed a policy distillation framework to obtain an interpretable policy from one that has been obtained through reinforcement learning. Since RL-based controllers can achieve high performance, but are too black-box to run on an actual wind farm, our method bridges the gap between theoretical research and real-world deployment. To evaluate our method, we use SCADA data from Norther, a Belgian offshore wind farm, together with an imbalance settlement simulator that combines historical data with the controller's actions. The distilled policy has been shown to achieve a significant reduction in imbalance costs, with a reduction of $82\%$ compared to $93\%$ using SAC. However, in contrast to SAC, this policy is interpretable since it is based on a transparent model and its control actions can be justified to the stakeholders. Furthermore, it outperforms other interpretable control strategies such as a price threshold controller, often used in industry, or a distilled policy using a decision tree.

\section*{Acknowledgments}

The authors gratefully acknowledge the financial support of the Flemish Government through the Flanders AI Research Program, the INSPIRE project (project no. SBEP2023-440) funded by the Sustainable Blue Economy Partnership and the Blue Cluster project called CORE funded by Flanders Innovation \& Entrepreneurship (VLAIO).

\bibliography{aaai2027}

\appendix

\section{Appendix}

\subsection{Price threshold controller}
\label{app:price_threshold}

The price threshold controller decides the power setpoint $x_t$ based on the latest published imbalance price estimate $\hat{\lambda}^{\mathrm{B}}_t$:
\begin{equation}
    x_t = \begin{cases}
        \overline{P}_t\quad\text{ if }\hat{\lambda}^{\mathrm{B}}_t>\overline{\lambda}^\mathrm{B}\\
        \underline{P}\quad\ \text{ if }\hat{\lambda}^{\mathrm{B}}_t<\underline{\lambda}^\mathrm{B}\\
        v_t\ \ \quad\text{else},\\
    \end{cases}
\end{equation}
where $\underline{P}$ denotes the minimum power setpoint and the price thresholds $\overline{\lambda}^\mathrm{B}$ and $\underline{\lambda}^\mathrm{B}$ are calibrated using historical price information following the procedure of \cite{bruninx2026herd}. In this work, they are set to $\underline{\lambda}^\mathrm{B}=-186$ €/MWh and $\overline{\lambda}^\mathrm{B}=230$ €/MWh.

\subsection{Soft decision trees}
\label{app:sdt}

We have implemented soft decision trees \citep[SDT;][]{frosst2017distilling} as an additional method for policy distillation following \cite{coppens2019distilling, gokhale2024distill2explain}. However, since we are performing regression between the policy of the student and the teacher, we had to use a mean squared error loss function rather than a cross entropy loss or a KL divergence loss. 

After training the model, predictions can be obtained in two ways:
\begin{enumerate}
    \item Take the prediction of the leaf with the maximum path probability (Method = Max)
    \item Averaging all leaf predictions by their corresponding path probabilities (Method = Mean)
\end{enumerate}
In the original SDT paper by \citet{frosst2017distilling}, the authors argue that the former makes the model explainable, whereas the latter only achieves ``marginally better test accuracy'' at ``an exponential increase in complexity of the explanation''. We agree with the authors that the second approach does not correspond to the definition of a transparent model employed in this paper. However, we find that, in our regression setting, the explainable predictions result in a significant drop in accuracy, as shown in Table \ref{table:sdt}, attaining even worse performance than a simple decision tree.

\begin{table}[h!]
\caption{Performance evaluation metrics for the SDT models. The inference method is specified in the `Method' column (cf., below). Mean absolute error (MAE) and root mean squared error (RMSE) are normalized by the capacity of the wind farm. Lower scores indicate a better fit for MAE and RMSE, while higher scores are better for $R^2$.}
\label{table:sdt}
\centering
\begin{tabular}{lcccc}
    \toprule
    & Method & MAE & RMSE & $R^2$ \\
    \midrule
    SDT (depth=2) & Mean & 0.20 & 0.26 & 0.50 \\
    SDT (depth=3) & Mean & 0.18 & 0.24 & 0.56 \\
    SDT (depth=2) & Max & 0.26 & 0.33 & 0.18 \\
    SDT (depth=3) & Max & 0.24 & 0.31 & 0.27 \\
    \bottomrule
\end{tabular}
\end{table}

\subsection{Expert-based EBM}
\label{app:expert_based}

The expert-based model only includes one single univariate effect of the imbalance price estimate $\hat{\lambda}^{\mathrm{B}}_t$ such that it closely resembles the price threshold controller. Furthermore, it includes six interaction effects:
\begin{enumerate}
    \item The interaction between the imbalance price estimate $\hat{\lambda}^{\mathrm{B}}_t$ and the system imbalance forecast $\hat{SI}^{\mathrm{ISP}}_t$
    \item The interaction between the imbalance price estimate $\hat{\lambda}^{\mathrm{B}}_t$ and the accumulated position during the settlement period $\Delta E_t$
    \item The interaction between the estimated power forecast error $\hat{\varepsilon}_t$ and the system imbalance forecast $\hat{SI}^{\mathrm{ISP}}_t$
    \item The interaction between the system imbalance forecast $\hat{SI}^{\mathrm{ISP}}_t$ and the minute of the settlement period $t^{\mathrm{ISP}}$
    \item The interaction between the imbalance price estimate $\hat{\lambda}^{\mathrm{B}}_t$ and the minute of the settlement period $t^{\mathrm{ISP}}$
    \item The interaction between the accumulated position during the settlement period $\Delta E_t$ and the estimated power forecast error $\hat{\varepsilon}_t$
\end{enumerate}

\end{document}